\documentclass{article}
\usepackage{spconf,amsmath,graphicx}
\usepackage{cite}
\usepackage{multirow}
\usepackage{amssymb,amsfonts}
\usepackage{algorithmic}
\usepackage{graphicx}
\usepackage{textcomp}
\usepackage{xcolor}
\usepackage{float}                  
\usepackage{subfig}
\usepackage{enumitem}
\setlist{nosep, leftmargin=14pt}

\usepackage{mwe} 

\pdfoutput = 1
\title{Semi-supervised Medical Image Segmentation via Query Distribution Consistency}
\name{Rong Wu$^{1,3,\dagger}$\thanks{$^\dagger$Corresponding author: Rong Wu (rw2867@nyu.edu)}, Dehua Li$^{2,3}$, Cong Zhang$^{3}$ }

\address{
$^{1}$ School of Global Public Health, New York University, New York, NY, USA \\
     $^{2}$ State Key Laboratory of Respiratory Disease, Guangzhou Institute of Respiratory Diseases, \\
     First Affiliated Hospital, Guangzhou Medical University, Guangzhou, China\\
    $^{3}$ DecisionLinnc Dev Group, Hangzhou, China\\
    }
\begin{document}
%
\maketitle
\begin{abstract}
Semi-supervised learning is increasingly popular in medical image segmentation due to its ability to leverage large amounts of unlabeled data to extract additional information. However, most existing semi-supervised segmentation methods focus only on extracting information from unlabeled data.
In this paper, we propose a novel Dual KMax UX-Net framework that leverages labeled data to guide the extraction of information from unlabeled data. Our approach is based on a mutual learning strategy that incorporates two modules: 3D UX-Net as our backbone meta-architecture and KMax decoder to enhance the segmentation performance.
Extensive experiments on the Atrial Segmentation Challenge dataset have shown that our method can significantly improve performance by merging unlabeled data. Meanwhile, our framework outperforms state-of-the-art semi-supervised learning methods on 10\% and 20\% labeled settings. Code located at: https://github.com/Rows21/DK-UXNet.
\end{abstract}
\section{Introduction}
\label{sec:intro}
As one of the most complicated tasks in computer vision, automated biomedical image segmentation, which aims to act like experienced physicians to identify types of tumors and delineate different sub-regions of organs on medical images such as MRI and CT scans, plays an important role in disease diagnosis \cite{basak2023pseudo, wang2023mcf, bian2023artificial}. Robust and precise segmentation of organs or lesions from medical images is an essential task in many clinical applications of diagnosis and treatment planning. With the recent emergence of neural networks, supervised deep learning methods have achieved state-of-the-art (SOTA) performance in multiple medical image segmentation tasks \cite{chen2017deeplab,milletari2016v,ronneberger2015u}. Many high-performance medical image segmentation methods rely heavily on collecting and annotating training data. However, for real-world medical images, annotations are often expensive to acquire as both expertise and time are needed to produce accurate annotations, especially in 3D volumetric images. 

Semi-supervised learning is a promising approach for processing images with limited supervised data. In recent years, semi-supervised methods based on consistency regularization \cite{sohn2020fixmatch,xie2020unsupervised} have attracted the attention of researchers and are one of the mainstream technologies. However, the role of labeled data has been largely ignored, and most semi-supervised learning algorithms consider labeled data as the initial step of the training pipeline or a means to ensure convergence \cite{kwon2022semi,ouali2020semi,luo2022semi}. Recently, the use of labeled data to directly guide the extraction of information from unlabeled data has attracted large attention \cite{wu2023querying,gao2023correlation}. In the field of semi-supervised medical image segmentation, there are shared features between labeled and unlabeled data, as well as out-of-distribution (OOD) cases, which have greater intuitiveness and guidance for algorithms. Typically, partially labeled clinical datasets exhibit similar foreground features, including comparable textures, shapes, and appearances between different samples.

In previous work, ConvNeXt \cite{liu2022convnet} refined the architectural insights of Vision Transformers \cite{dosovitskiy2020image} and Swin Transformers \cite{liu2021swin} into convolutional architectures. The ConvNeXt block inherits many significant structures from Transformers, aiming to limit computational costs while expanding the network, demonstrating performance improvements compared to standard ResNets \cite{he2016deep}. 
On the other hand, Mask Transformers attempts to enhance CNN-based backbone networks through independent Transformer blocks. MaX-Deeplab\cite{wang2021max} interprets object queries in DETR \cite{carion2020end} as memory-encoded queries for end-to-end panoramic segmentation. MaxQuery\cite{yuan2023devil} and KMaX-Deeplab \cite{yu2022k} proposed interpreting queries as cluster centers and adding regulatory constraints for learning the clustering representation of queries.

Inspired by the works mentioned above, we introduce a novel $\mathbf{D}$ual-$\mathbf{K}$Max $\mathbf{UX}$-Net (DKUXNet) for semi-supervised medical image segmentation.
In our work, we leverage these strengths by adopting the general design of 3D UX-Net \cite{lee20223d} and kMax decoder as our backbone meta-architecture.
Adopted from \cite{yu2022k}, we propose the cluster classification consistency to regularize a specific number of object queries related to background, organs, and tumors.
The main contributions of our work are as follows: 
(1) We have developed a new semi-supervised segmentation model that divides images into three categories: background, organ, and tumor and calculates and updates the distance of the cluster center. Its performance is similar to state-of-the-art fully supervised models, but it utilizes 20\% of training data. 
(2) We propose a new strategy that utilizes the consistency loss of query distribution and segmentation outputs for backpropagation calculations to enhance image consistency.

\section{Method}
\subsection{Overview}
\begin{figure}[htbp]
\centerline{\includegraphics[width=\columnwidth]{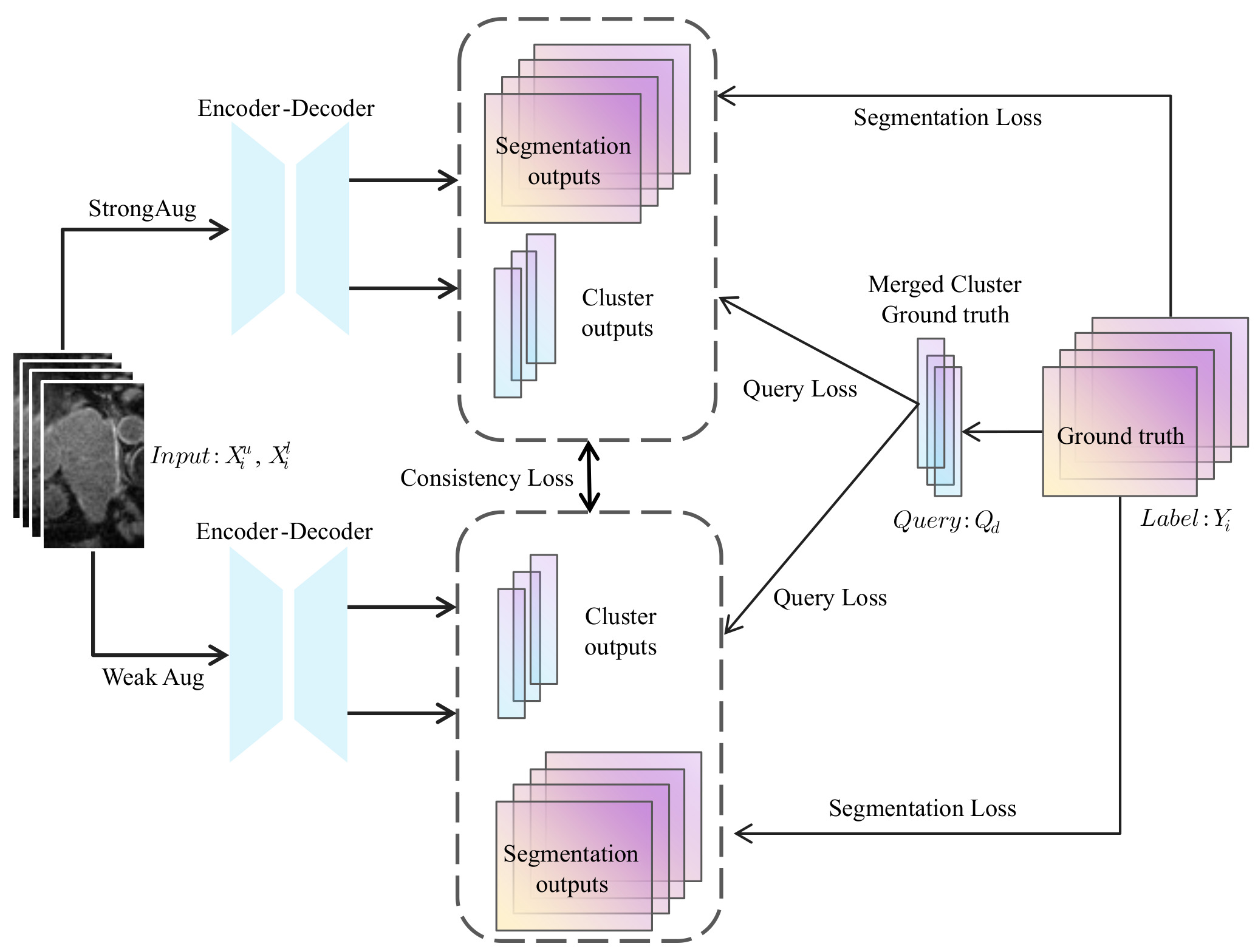}}
\caption{Overall workflow of our proposed method. Our proposed dual KMax-based contrastive learning strategy (details can be found in Section \ref{unet} and Section \ref{kmean}).}
\label{fig1}
\end{figure}
As shown in Fig. \ref{fig1}, we build our query-based segmentation network by a fully ConvNeXt backbone and a transformer-based module. Given the $N_L$ samples labeled set $(X_i^l, Y_i), i \in (1,N_L)$ and $N_U$ samples unlabeled set $X_i^u, i \in (1,N_U)$ as the input, we perform strong and weak augmentations on the same input image, which transformed to strong and weak augmented data. The proposed segmentation network aims to learn information from both $X^u_{is}, X^l_{is}$ and $X^u_{iw}, X^l_{iw}$, which outputs the predicted segmentation $Y_i^{pred}$ and query distribution logits $Q_d$ (in Section \ref{kmean}) for consistency loss. Label $Y_i$ and ground-truth query distribution $Q_{d}$ is used to supervise the cluster outputs (in Section \ref{kmean}), respectively.

\subsection{Basic Structures}
\subsubsection{ConvNeXt Based Segmentation Architecture}\label{unet}
\begin{figure}[htbp]
\centerline{\includegraphics[width=\columnwidth]{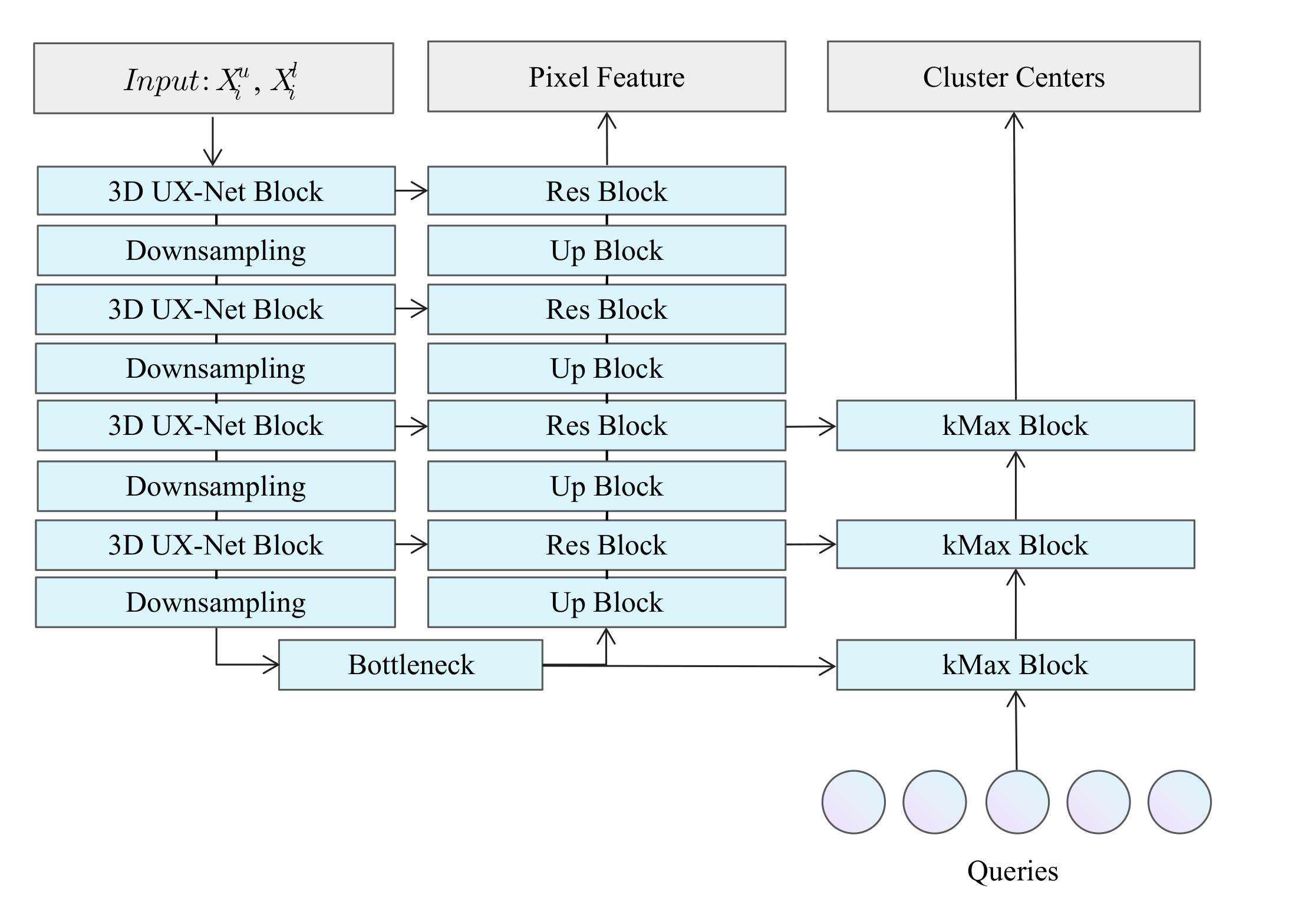}}
\caption{The meta-architecture of the backbone network consists of three components: ConvNeXt encoder, CNN-based decoder, and kMaX decoder.}
\label{fig}
\end{figure}
In our work, we utilize these advantages by adopting the general design of 3D UX-Net \cite{lee20223d} as the building block in our backbone. 
The ConvNeXt back includes transformer encoders to enhance the pixel features, and upsampling layers to generate higher-resolution features. We use four 3D UX-Net blocks and four Downsampling blocks as the depth-wise convolution encoder. 
The multi-scale outputs from each stage in the encoder are connected to a ConvNet-based decoder via skip connections. Specifically, we extract the outputs for stage $i$ ($i\in 2,3,4$) in the encoder and further deliver the outputs into kMax decoder block (Section \ref{kmean}) for cluster center information learning.

\subsubsection{K-Means Cross-attention Algorithm}\label{kmean}
\begin{figure}[htbp]
\centerline{\includegraphics[width=\columnwidth]{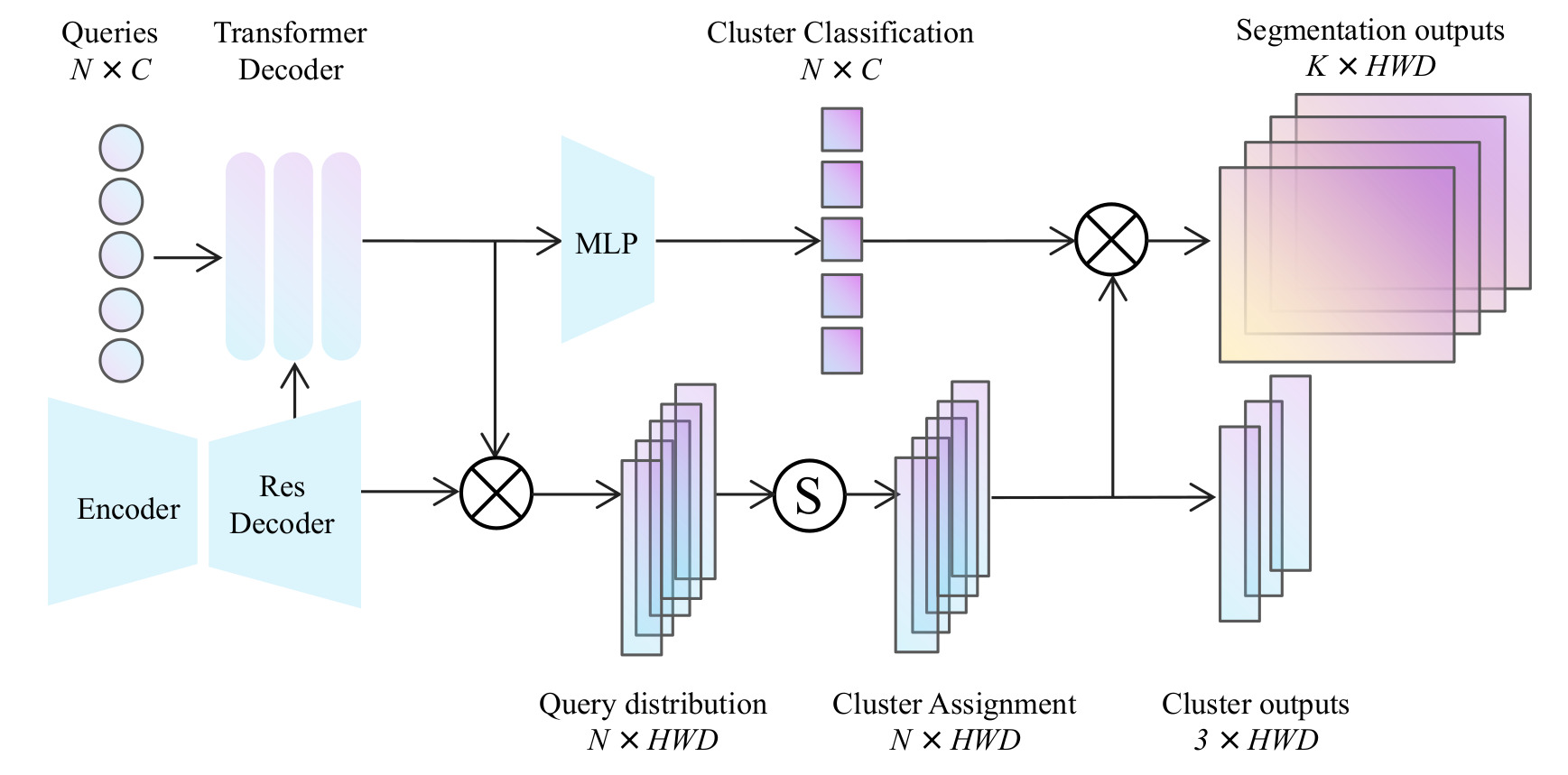}}
\caption{An illustration of kMaX UX-Net.}
\label{fig}
\end{figure}
In our transformer module, we create a set of object queries $\mathbf{C} \in \mathbb{R}^{N \times C}$ with $N$ classes and $C$ channels. The transformer objects use the extracted pixel feature outputs from pixel decoder and gradually updates the learnable object queries to interpretive picture features. The classical cross attention algorithm between object queries and per-pixel features was:
\begin{equation}
    \begin{split}
        \mathbf{C} = \mathbf{C} + \arg\max_N(\mathbf{Q}^c (\mathbf{K}^p)^T)\mathbf{V}^p, \label{eq_vit}
    \end{split}
\end{equation}
where the superscripts $c$ and $p$ represent query and pixel features, respectively.
Inspired by recent works on cluster analysis of mask transformers \cite{yu2022k,yuan2023devil}, we adopt cluster-wise argmax to substitute spatial-wise softmax in original cross-attention settings, and the new update algorithm work as follows:
\begin{equation} \label{eq_kmax}
    \begin{split}
        &\mathbf{A} = \arg\max_N(\mathbf{C} \times \mathbf{P}^T), \\ 
        &\mathbf{\hat{C}} = \mathbf{A} \times \mathbf{P},
    \end{split}
\end{equation}
where $\mathbf{C} \in \mathbb{R}^{N \times C}$, $\mathbf{P} \in \mathbb{R}^{HWD \times C}$, and $\mathbf{A} \in \mathbb{R}^{HWD \times N}$ refers to cluster center, pixel features and cluster assignments. 

\subsubsection{Postprocessing}
Adopted from \cite{yuan2023devil}, we construct this postprocessing to maximize the Dice score of the resulting segmentation. The pixel feature output $\mathbf{P}$ from pixel decoder is first produced with the predicted cluster center $\mathbf{C}$ to generate a query response $\mathbf{R}$. Then, we use softmax activation on query responses $\mathbf{R}$ to generate a mask prediction, which encourages the exclusiveness of cluster assignment. Secondly, the grouped pixels are classified under the guidance of cluster classification. We evaluate the cluster centers $\mathbf{C}$ via a multi-layer perceptron (MLP) to predict the K-class cluster classifications $\mathbf{C_k}\in \mathbb{R}^{N \times K}$. We then aggregate the cluster assignments $\mathbf{M}$ of grouped pixels and their classifications $\mathbf{C_k}$ for the final segmentation outputs.

\subsection{Segmentation Framework with Contrastive Loss}
We propose a novel Dual-Contrastive Loss ($\mathcal{L}_{dc}$) to measure the difference between to augmented sets. Previous works such as JCL \cite{cai2020joint} compute the expectation of the InfoNCE loss \cite{oord2018representation} over a distribution of positive samples only, for a given query. In mathematical terms, the InfoNCE loss is defined as follows:
\begin{equation}\label{qdcl}
    \begin{split}
        \mathcal{L}_{InfoNCE} = -\log\dfrac{\exp(sim(z_i,z_j)/\tau)}{\sum_{k \neq i}\exp(sim(z_i,z_k)/\tau)},
    \end{split}
\end{equation}
where $i$ and $j$ correspond to two data augmentations of the same original image, $z$ denotes network output, and $\tau$ denotes the temperature parameter \cite{chen2020simple}. Following SimCLR \cite{chen2020simple}, the similarity function function used here is the cosine similarity: $sim(x,y) = \dfrac{x^Ty}{\left\lVert x\right\rVert \left\lVert y\right\rVert}$.

In our case, assume that $N_L$ labeled samples $(X_i^l,Y_i), i \in (1,N_L)$ and $N_U$ unlabeled samples $x_i^u, i \in (1,N_U)$ are put into the model, where $X_i \in \mathbb{R}^{HWD}$ and $Y_i \in \mathbb{R}^{K \times HWD}$ stands for input volume and annotated $K$ classes ground-truth. $\mathcal{L}_{dc}$ includes InfoNCE loss at two levels: query distribution $\mathcal{L}_{qdc}$ and predicted segmentation output $\mathcal{L}_{segc}$, described as follows:
\begin{equation}\label{qdc}
    \begin{split}
        &\mathcal{L}_{segc} = -\dfrac{1}{HWD}\log\dfrac{\exp(sim(X_i,X_j)/\tau)}{\sum_{k \neq i}\exp(sim(X_i,X_k)/\tau)},\\
        &\mathcal{L}_{qdc} = -\dfrac{1}{HWD}\log\dfrac{\exp(sim(Q_i,Q_j)/\tau)}{\sum_{k \neq i}\exp(sim(Q_i,Q_k)/\tau)},
    \end{split}
\end{equation}
where. For supervised learning, we combine the cross entropy and dice loss between the final outputs and the ground truth as the segmentation loss in the equation, which is $\mathcal{L}_{sup} = \mathcal{L}_{ce} + \mathcal{L}_{dc}$. The final loss function $\mathcal{L}$ is a combination of supervised loss and Dual-Contrastive Loss $\mathcal{L}_{dc}$ with a balance weight $\lambda$, formulated as,
\begin{equation}\label{finalloss}
    \begin{split}
        &\mathcal{L}_{l} = \mathcal{L}_{sup} + \lambda\mathcal{L}_{dc}, \\
        &\mathcal{L}_{u} = \lambda\mathcal{L}_{dc}, \\ 
        &\mathcal{L}_{dc} = \mathcal{L}_{segc} + \mathcal{L}_{qdc}, \\
    \end{split}
\end{equation}
where $\mathcal{L}_{l}$ denotes the loss function for labeled samples and $\mathcal{L}_{u}$ for unlabeled samples.

\section{Experiments}
\subsection{Dataset and Experiment Setting}
To evaluate the proposed method, we apply our algorithm on Left Atrial (LA) dataset \cite{xiong2021global} from the 2018 Atrial Segmentation Challenge. The dataset consists of 100 3D gadolinium-enhanced MR images (GE-MRIs) and their ground truth, with a resolution of $0.625 \times 0.625 \times 0.625$mm. Following \cite{yu2019uncertainty}, we use 80 scans for training and 20 scans for validation and apply the same preprocessing methods. All scans are centered at the heart region cropped accordingly, and then normalized to zero mean and unit variance. In this work, we report the performance of all methods trained with three different settings of 5\%/10\%/20\% labeled images, which is the typical semi-supervised learning experimental setting \cite{xia20203d}. 

\subsection{Implementation Details and Evaluation Metrics}
Our DK-Unet model is implemented in Pytorch 1.12.1 and trained on four NVIDIA P100 GPUs with a batch size of 4, which cropped two training samples and made strong and weak augmentations. We conducted all experiments with fixed random seeds and 4000 epochs. The raw LA training data for each case are randomly cropped to $112 \times 112 \times 80$ voxels following [23]. For the optimization, we use the AdamW optimizer with an initial learning rate of 0.0001. Results are evaluated on four metrics: Dice, Jaccard Index, 95\% Hausdorff Distance (95HD), and Average Surface Distance (ASD). To ensure a fair comparison, we perform all experiments on the same machine and report the mean results from the final iteration.

\subsection{Main Results}
We conducted experiments on different percentages of labeled data and compared their performance with the corresponding data trained in a fully supervised manner in Table \ref{tab1}. As shown in the first two rows of Table \ref{tab1}, our method can mine distinctive features through the fully supervised method, thereby exceeding the results generated by V-Net. We compared DKUXNet with other baselines on LA. Our framework shows optimal performance in most metrics. Specifically, with only 5\% labeled data, DKUXNet achieved 85.96\% of the dice score. DKUXNet also achieved a 90.41\% dice score with only 10\% labeled data. When the labeled data volume increased to 20\%, the results obtained by this model were comparable to those of V-Net trained in 100\% labeled data, and compared to the 90.98\% score of the upper bound model, Dice scored 91.70\%.
\begin{table}[htbp]
\centering
\caption{Comparison with SOTA methods on the LA dataset.}
\setlength\tabcolsep{3.05pt}
\begin{center}
\begin{tabular*}{\linewidth}{l|c|cccc}
\hline
\multirow{2}{*}{Method} & \multicolumn{1}{c|}{\multirow{2}{*}{\begin{tabular}[c]{@{}c@{}}Labeled\\ Scans\end{tabular}}} & \multicolumn{4}{c}{Metrics} \\ \cline{3-6} 
                        &   & Dice$\uparrow$ & Jaccard$\uparrow$ & 95HD$\downarrow$ & ASD$\downarrow$\\ \hline
V-Net                   & \multirow{2}{*}{80} & 90.98 & 83.61 & 8.58 & 2.10 \\ 
ours                   &                                          & 92.62 & 86.32 & 4.62 & 1.28  \\ \cline{1-6} 
SASSNet\cite{li2020shape}& \multirow{8}{*}{4(5\%)}   & 78.07 & 65.03 & 29.17 & 8.63 \\
DTC\cite{luo2021semi} &                              & 79.61 & 67.00 & 25.54 & 7.20 \\
MC-Net\cite{wu2021semi} &                              & 80.14 & 67.88 & 24.08 & 7.18 \\
URPC\cite{luo2022semi} &                               & 80.92 & 68.90 & 17.25 & 2.76 \\
SS-Net\cite{wu2022exploring} &                         & 80.75 & 68.54 & 19.81 & 4.98 \\
MC-Net+\cite{wu2022mutual} &                           & 83.33 & 71.79 & 15.70 & 4.33 \\
CAML\cite{gao2023correlation}&                         & $\mathbf{87.34}$ & $\mathbf{77.65}$ & $\mathbf{9.76}$ & $\mathbf{2.49}$ \\
ours                    &                              & 85.96 & 75.91 & 11.72 & 2.64 \\ \cline{1-6} 
SASSNet\cite{li2020shape}& \multirow{8}{*}{8(10\%)}   & 85.71 & 75.35 & 14.74 & 4.00 \\
DTC\cite{luo2021semi}&                                 & 84.55 & 73.91 & 13.80 & 3.69 \\
MC-Net\cite{wu2021semi}&                               & 86.87 & 78.49 & 11.17 & 2.18 \\
URPC\cite{luo2022semi}&                                & 83.37 & 71.99 & 17.91 & 4.41 \\
SS-Net\cite{wu2022exploring} &                         & 86.56 & 76.61 & 12.76 & 3.02 \\
MC-Net+\cite{wu2022mutual}&                            & 87.68 & 78.27 & 10.35 & 1.85 \\
CAML\cite{gao2023correlation}&                         & 89.62 & 81.28 & 8.76 & 2.02 \\
ours                    &                              & $\mathbf{90.41}$ & $\mathbf{82.69}$ & $\mathbf{7.32}$ & $\mathbf{1.71}$ \\ \cline{1-6} 
SASSNet\cite{li2020shape}& \multirow{8}{*}{16(20\%)}   & 88.11 & 79.08 & 12.31 & 3.27 \\
DTC\cite{luo2021semi}&                                 & 87.79 & 78.52 & 10.29 & 2.50 \\
MC-Net\cite{wu2021semi} &                              & 90.43 & 82.69 & 6.52 & 1.66 \\
URPC\cite{luo2022semi}&                                & 87.68 & 78.36 & 14.39 & 3.52 \\
SS-Net\cite{wu2022exploring}&                          & 88.19 & 79.21 & 8.12 & 2.20 \\
MC-Net+\cite{wu2022mutual}&                            & 90.60 & 82.93 & 6.27 & 1.58 \\
CAML\cite{gao2023correlation}&                         & 90.78 & 83.19 & 6.11 & 1.68 \\
ours                    &                              & $\mathbf{91.70}$ & $\mathbf{84.82}$ & $\mathbf{5.81}$ & $\mathbf{1.62}$ \\ \cline{1-6} 
\end{tabular*}
\label{tab1}
\end{center}
\end{table}
\subsection{Ablation Studies}

\begin{table}[htbp]
\caption{Ablation study of our DKUXNet on the LA dataset under 10\% labeled scenario.}
\begin{center}
\begin{tabular}{ccc|cccc}
\hline
\multicolumn{3}{c|}{Components}                  & \multicolumn{4}{c}{Metrics}          \\ \cline{1-7} 
\multicolumn{1}{c}{$\mathcal{L}_{seg}$} & \multicolumn{1}{c}{$\mathcal{L}_{qdc}$} & \multicolumn{1}{c|}{$\mathcal{L}_{segc}$} & \multicolumn{1}{c}{Dice$\uparrow$} & \multicolumn{1}{c}{Jaccard$\uparrow$} & \multicolumn{1}{c}{95HD$\downarrow$} & ASD$\downarrow$ \\ \hline
\checkmark &            &           & 77.73 & 64.42 & 16.86 & 3.92 \\
\checkmark & \checkmark &       & 82.12 & 70.16 & 23.63 & 5.99 \\
\checkmark &            & \checkmark & 85.52 & 75.49 & 14.90 & 3.73 \\
\checkmark & \checkmark & \checkmark & 90.41 & 82.69 & 7.32 & 1.71 \\ \hline

\end{tabular}
\label{Ablation}
\end{center}
\end{table}

We conducted ablation experiments to verify the effectiveness of the key components in our proposed model. To investigate the individual impact of different tasks, we first only use labeled images for training and analyze how the dual-task consistency performs when only labeled images are used. As shown in Table 2, dual-contrastive loss substantially improves segmentation performance when labeled data is limited. 
The performance of these variants is listed in Table \ref{Ablation}.

\begin{figure}[htbp]
\centerline{\includegraphics[width = 1\columnwidth]{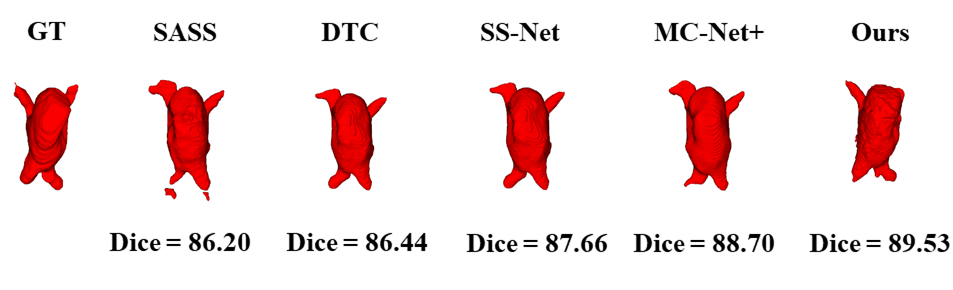}}
\caption{3D Visualization of different ablation studies for LA segmentation. GT: ground truth. (best viewed in color)}
\label{fig_res}
\end{figure}

Fig. \ref{fig_res} shows our visualization results compared with other methods under 10\% labeled scenario. Our framework outperforms state-of-the-art semi-supervised learning methods on 10\% and 20\% labeled settings. 

\section{CONCLUSION}
In this study, we developed a semi-supervised learning framework based on consistency loss with query distribution and segmentation outputs. Our key idea is that (query) cluster assignment should be considered for semi-supervised learning. The experimental results indicate that our method fully achieves performance similar to the state-of-the-art method. However, the performance of the proposed method is not as outstanding at 5\% labeled setting, we should further develop novel consistency loss for information transfer between unlabeled data and label data and test on more complicated segmentation datasets. We believe that the proposed method has the good potential to boost further the option of medical image segmentation in designing various clinical applications with minimum labeled data. 

\section{COMPLIANCE WITH ETHICAL STANDARDS}
This is a numerical simulation study for which NO ethical approval was required.
\bibliographystyle{IEEEbib}

\end{document}